\documentclass[10pt,twocolumn,letterpaper]{article}

\usepackage{cvpr}      

\input{preamble}

\definecolor{cvprblue}{rgb}{0.21,0.49,0.74}
\usepackage[pagebackref,breaklinks,colorlinks,allcolors=cvprblue]{hyperref}

\def\paperID{3} 
\def\confName{CVPR}
\def\confYear{2026}

\title{SCR-Guided Difficulty-Aware Optimization for Infrared Small Target Detection}

\author{Yunus Sevim\\
Aselsan\\
Ankara, Turkey\\
{\tt\small ysevim@aselsan.com}
\and
Behçet Uğur Töreyin\\
Istanbul Technical University\\
Istanbul, Turkey\\
{\tt\small toreyin@itu.edu.tr}
}

\begin{document}
\maketitle
\begin{abstract}
Infrared small target detection remains challenging due to severe background clutter,
low contrast, and weak spatial responses where geometric overlap alone is insufficient
to characterize detection quality.
In this work, we propose REEM (Reweighted Explicit-visibility Enhanced Modulation), a lightweight SCR-guided difficulty-aware optimization
framework that incorporates Signal-to-Clutter Ratio (SCR) as a physically meaningful
visibility prior during training.
Instead of modifying the network architecture or directly optimizing SCR,
REEM computes a ground-truth local SCR from the input image and applies a
differentiable modulation to the soft-IoU learning signal, emphasizing low-visibility
targets while preserving stable optimization and identical inference behavior.
REEM is integrated into a U-Net-based MSHNet without introducing additional parameters,
architectural modifications, or inference-time overhead.
Extensive experiments demonstrate consistent improvements over the baseline,
achieving higher IoU and detection probability (Pd) together with substantially
reduced false alarms (FA), particularly under challenging low-visibility conditions.
These results suggest that SCR-guided difficulty-aware optimization provides an
effective and physically grounded complement to conventional overlap-based objectives
for infrared small target detection.
The code is available at \url{https://github.com/yall-in-one/Reemm}.
\end{abstract}    
\section{Introduction}
\label{sec:intro}

Infrared small target detection (IRSTD) remains one of the core 
challenges in infrared search and tracking systems, where reliable 
perception under adverse sensing conditions is critical~\cite{Zhao2022SurveyIRSTD}.
Infrared sensors are increasingly deployed in safety-critical 
pipelines alongside visible-range cameras, particularly when 
illumination, weather, or operational constraints degrade the effectiveness of visible-range small 
object detection methods~\cite{Koyun2022FocusDetect}.
IRSTD has broad applications in both civil and military domains, 
including traffic management~\cite{Zhang2021ReviewIRSTD}, maritime 
surveillance and rescue, navigation, early warning systems, and precise 
guidance~\cite{Zhang2022ISNet, DNANet2023}.
In contrast to generic object detection~\cite{Redmon2018YOLOv3, 
Zou2023ObjectDetectionSurvey}, which assumes rich appearance cues 
and sufficient spatial extent, IRSTD must cope with long sensing 
distances, heavy background clutter, sensor noise, and low 
signal-to-clutter ratios (SCR), where targets typically span only 
a few pixels with weak contrast~\cite{Gao2013InfraredPatch, 
Zhu2020TNLRS, Kim2015HighSpeedIRSTD}.

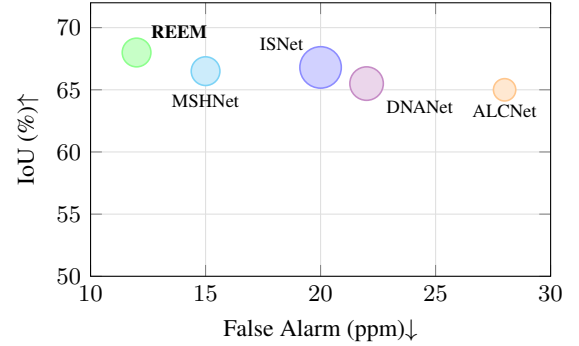
\begin{figure}[t]
\centering
\begin{tikzpicture}
\begin{axis}[
    width=0.92\linewidth,
    height=5.2cm,
    xmin=10, xmax=30,
    ymin=50, ymax=72,
    xlabel={False Alarm (ppm)$\downarrow$},
    ylabel={IoU (\%)$\uparrow$},
    ylabel style={rotate=0},
    grid=both,
    major grid style={gray!25},
    minor grid style={gray!12},
    tick label style={font=\small},
    label style={font=\small},
]

\def\B{6.0}

\addplot[
    only marks, mark=*,
    mark size=0.90*\B,
    fill=green!22, draw=green!45, line width=0.6pt
] coordinates {(12,68)};
\node[
    anchor=south west,
    font=\scriptsize\bfseries,
    xshift=2pt, yshift=2pt
] at (axis cs:12,68) {REEM};

\addplot[
    only marks, mark=*,
    mark size=0.90*\B,
    fill=cyan!18, draw=cyan!45, line width=0.6pt
] coordinates {(15,66.5)};
\node[
    anchor=north,
    font=\scriptsize,
    yshift=-5pt
] at (axis cs:15,66.5) {MSHNet};

\addplot[
    only marks, mark=*,
    mark size=1.05*\B,
    fill=violet!16, draw=violet!45, line width=0.6pt
] coordinates {(22,65.5)};
\node[
    anchor=north west,
    font=\scriptsize,
    xshift=4pt, yshift=-3pt
] at (axis cs:22,65.5) {DNANet};

\addplot[
    only marks, mark=*,
    mark size=0.70*\B,
    fill=orange!18, draw=orange!45, line width=0.6pt
] coordinates {(28,65)};
\node[
    anchor=north,
    font=\scriptsize,
    yshift=-2pt
] at (axis cs:28,65) {ALCNet};

\addplot[
    only marks, mark=*,
    mark size=1.30*\B,
    fill=blue!18, draw=blue!45, line width=0.6pt
] coordinates {(20,66.8)};
\node[
    anchor=south east,
    font=\scriptsize,
    xshift=-3pt, yshift=3pt
] at (axis cs:20,66.8) {ISNet};

\end{axis}
\end{tikzpicture}
\caption{
IoU--FA trade-off of representative deep learning-based 
infrared small-target detectors on IRSTD-1k (test).
The x-axis denotes the false alarm rate (FA) in parts per 
million (ppm), and the y-axis reports IoU (\%).
Circle area indicates relative model complexity (e.g., 
parameter count as a proxy).
REEM achieves higher IoU at lower false alarm rates compared 
to prior methods, suggesting a better accuracy-reliability 
trade-off.
}
\label{fig:tradeoff_irstd}
\end{figure}

Infrared small targets under long-range imaging conditions often appear 
as localized PSF-like intensity responses rather than spatially 
well-defined objects. Diffraction effects, limited sensor resolution, 
and atmospheric degradation cause targets to lack stable geometric 
structures, instead producing compact responses with ambiguous spatial 
extent and weak shape or texture cues~\cite{Dai2021AsymmetricContext}. 
Consequently, detection performance in IRSTD is often governed more by 
target--background separability than by precise geometric overlap, 
which limits the effectiveness of purely appearance-driven learning objectives.

Early IRSTD methods predominantly relied on model-driven formulations, including filtering-based approaches, local contrast enhancement, and low-rank background modeling~\cite{Deshpande1999MaxMedian, Gao2013InfraredPatch, Han2019TriLayerLCM}. While effective in constrained settings, these model-driven approaches rely heavily on handcrafted priors and are sensitive to hyper-parameter selection, often degrading under complex clutter and dynamic backgrounds~\cite{Yang2020Subspace, Ying2023MappingDegeneration}.

With the increasing demand for rapid response and early warning capabilities, single-frame infrared small target detection (SIRST) has attracted growing attention~\cite{Kim2015HighSpeedIRSTD}, particularly in scenarios where temporal continuity assumptions break down due to fast-moving platforms or targets, making single-frame spatial modeling a more reliable alternative~\cite{Gao2013InfraredPatch}. Recent advances in deep learning, especially encoder--decoder architectures such as U-Net, have enabled data-driven IRSTD methods that learn discriminative representations directly from annotated data, alleviating the reliance on handcrafted priors~\cite{Ronneberger2015UNet, Liu2018SNR, DNANet2023}. Notable progress has been achieved through specialized network architectures, including dense nested structures, multi-scale feature aggregation, attention mechanisms, contextual modeling, and more recently transformer- and state-space-inspired designs~\cite{DNANet2023, MSHNet2024, Zhang2025SAIST, Yuan2024SCTransNet, Chen2024MiMISTD}.

Despite these architectural advances, most existing deep learning-based
IRSTD methods continue to rely on generic supervision strategies,
such as Intersection-over-Union (IoU) loss and Dice loss, originally
designed for objects with well-defined spatial extents~\cite{Sudre2017Dice}.
However, in regimes where infrared targets exhibit weak visibility,
localized intensity responses, or PSF-like characteristics,
overlap-based losses may correlate only weakly with detection reliability.
Targets with substantially different detectability can produce similar
overlap scores, leading to misaligned optimization signals and
suboptimal performance, particularly for dim targets embedded in
complex background clutter~\cite{Dai2021AsymmetricContext, Gao2013InfraredPatch}.

MSHNet recently showed that meaningful performance gains in IRSTD 
do not necessarily require more complex architectures~\cite{MSHNet2024}. 
Instead, revisiting the loss function itself can guide feature learning 
more effectively while keeping the model compact and deployment-friendly. We build on this insight and propose REEM (Reweighted Explicit-visibility Enhanced Modulation), which reweights the 
training signal according to ground-truth local SCR. Low-visibility 
targets receive higher gradient emphasis, while high-SCR samples 
are down-weighted, directly addressing the mismatch between 
overlap-based losses and true detection difficulty. REEM requires 
no architectural changes and introduces no inference-time overhead, 
making it straightforward to apply on top of existing IRSTD frameworks.
\section{Related Work}
\label{sec:related}
This section reviews three areas directly relevant to the proposed 
method: single-frame IRSTD approaches, loss function design in IRSTD, 
and signal-based measures used to characterize target visibility.

\subsection{Single-Frame Infrared Small Target Detection}

Single-frame infrared small target detection (IRSTD) has been extensively studied due to its practical importance and inherent difficulty. Early IRSTD approaches mainly relied on handcrafted priors, including filtering-based methods~\cite{Deshpande1999MaxMedian}, local-contrast-based techniques~\cite{Han2019TriLayerLCM}, and low-rank modeling frameworks~\cite{Gao2013InfraredPatch, Zhu2020TNLRS, Yang2020Subspace}. Although effective in constrained scenarios, these methods are highly sensitive to parameter tuning and often exhibit limited robustness under complex and cluttered backgrounds.

With the development of deep learning, convolutional neural network (CNN)-based methods have substantially advanced single-frame IRSTD. Early deep models emphasized background suppression and local contrast enhancement through asymmetric contextual modulation and attention-guided learning~\cite{Dai2021AsymmetricContext, Zhao2022SurveyIRSTD}. These studies highlighted that suppressing background clutter while preserving weak target responses is essential for reliable infrared perception under challenging imaging conditions.

Subsequent works focused on architectural improvements to maintain 
small targets in deep representations. Shape-aware supervision was 
introduced to exploit geometric structure of small targets for improved 
localization~\cite{Zhang2022ISNet}. Dense nested encoder--decoder 
structures and attention-enhanced U-shaped networks were subsequently 
introduced to alleviate target disappearance in deep layers and 
strengthen multi-scale contextual modeling~\cite{DNANet2023, 
Ronneberger2015UNet}.

More recently, transformer-based and hybrid CNN--Transformer architectures have been explored to capture long-range dependencies and global contextual information, which is particularly beneficial for perception tasks beyond the visible spectrum where spatial cues are sparse and ambiguous~\cite{Chen2022IRSTFormer, Yuan2024SCTransNet}. In parallel, alternative supervision strategies and efficiency-oriented designs have also been investigated. Adversarial learning frameworks explicitly modeled the trade-off between missed detections and false alarms~\cite{Wang2019MDvsFA}, while single-point supervision and label-evolution schemes reduced annotation cost without sacrificing detection accuracy~\cite{Ying2023MappingDegeneration}. Lightweight and real-time detectors were proposed to balance detection performance with computational efficiency~\cite{Zhao2019TBCNet}.

Although these advances have significantly improved detection performance, most existing single-frame IRSTD methods predominantly emphasize architectural design and feature representation. Physically meaningful signal characteristics, which fundamentally govern infrared target detectability, remain relatively underexplored within the learning objectives of current deep models.

\subsection{Geometry-Driven Loss Functions in IRSTD}

Loss design plays a central role in deep learning-based infrared small target detection (IRSTD) by shaping how models interpret target–background discrepancies during optimization.
Most existing approaches rely on pixel-wise or geometry-driven objectives, including binary cross-entropy (BCE), Dice loss~\cite{Sudre2017Dice}, and Intersection-over-Union (IoU), which primarily measure spatial overlap between predictions and ground truth.
While effective for general segmentation tasks, such losses remain largely insensitive to variations in target visibility and signal strength, which are critical factors in infrared imagery where targets often appear as faint localized responses.

To improve detection robustness, several studies have explored 
alternative supervision strategies that remain largely grounded 
in geometric or appearance-driven paradigms. Adversarial learning 
frameworks explicitly model the trade-off between missed detections 
and false alarms through competitive optimization~\cite{Wang2019MDvsFA}, 
while context- or edge-aware mechanisms aim to enhance boundary 
sensitivity and local contrast modeling for small 
targets~\cite{Dai2021AsymmetricContext, Zhang2022ISNet}. 
Geometry-aware extensions, including Generalized IoU 
(GIoU)~\cite{Rezatofighi2019GIoU} and Complete IoU 
(CIoU)~\cite{Zheng2020CIoU}, further improve localization by 
incorporating geometric constraints such as scale and distance. 
Similarly, scale- and location-sensitive objectives have been 
introduced to address ambiguities caused by small target size 
variations~\cite{MSHNet2024}. Despite these advances, these 
geometry-driven approaches share a common limitation: they optimize 
spatial correspondence without accounting for visibility-related 
difficulty, leaving physically meaningful signal characteristics largely 
unexploited. This limitation motivates the exploration of signal-aware 
supervision mechanisms that incorporate visibility cues directly into 
the learning objective, forming the basis of the proposed SCR-guided 
optimization framework.

\subsection{Signal-Based Measures in IRSTD}

In infrared imaging, target detectability is fundamentally influenced 
by the strength of the target signal relative to background clutter 
and noise. Various signal-based measures have been explored in the 
IRSTD literature to characterize this relationship.

Signal-to-Noise Ratio (SNR) evaluates the contrast between the target 
signal and sensor noise and is commonly used to assess imaging 
quality~\cite{Liu2018SNR}. While useful in noise-dominated settings, 
SNR does not capture the effect of structured background clutter, 
which is the dominant source of difficulty in most infrared scenarios. Contrast-to-Noise Ratio (CNR) jointly considers contrast magnitude 
and noise variance to evaluate target 
separability~\cite{Bushberg2011MedicalImaging}. While widely used 
in medical and scientific imaging, its application in IRSTD remains 
limited, as background clutter rather than noise variance typically 
governs detection difficulty in infrared scenarios.

Signal-to-Clutter Ratio (SCR) directly measures the relative strength 
of the target signal against surrounding background 
structures~\cite{Gao2013InfraredPatch}, making it the most informative 
indicator of target visibility in cluttered scenes. Unlike SNR and CNR, 
SCR explicitly quantifies target-background separability under realistic 
infrared conditions, and has been widely adopted to characterize scene 
difficulty and analyze robustness in low-contrast infrared 
imagery~\cite{Zhao2022SurveyIRSTD}. This property makes SCR a natural 
candidate for integration into the learning objective, forming the core 
motivation of the proposed REEM framework.

Traditional IRSTD pipelines further employ local contrast descriptors 
and clutter-related statistics to enhance saliency or evaluate 
background suppression~\cite{Deshpande1999MaxMedian, 
Han2019TriLayerLCM, Gao2013InfraredPatch, Zhu2020TNLRS}, yet 
these measures remain confined to heuristic enhancement rather than 
end-to-end optimization, underscoring the potential value of 
signal-aware supervision strategies that better reflect the physical 
properties of infrared target visibility.

\section{Method}
\label{sec:method}

This section presents REEM, a lightweight SCR-guided loss reweighting 
framework for infrared small target detection. We first describe the 
baseline loss functions used in IRSTD, then introduce the proposed 
SCR-guided reweighting strategy and its integration into the MSHNet 
training pipeline.

\subsection{Overview of REEM}
\label{sec:reem_overview}

Infrared small target detection remains challenging due to severe 
background clutter, low contrast, and weak spatial responses where 
geometric overlap alone is often insufficient to characterize detection 
quality. Instead of introducing new network architectures, we propose 
\textit{REEM}, a lightweight SCR-guided difficulty-aware optimization 
framework that reshapes the training objective through physically 
meaningful visibility cues.

The key idea of REEM is to incorporate the Signal-to-Clutter Ratio 
(SCR) as a visibility prior during training. REEM computes a 
ground-truth local SCR from the input image and uses it to adaptively 
reweight the learning signal through a differentiable weighting 
function, increasing the contribution of low-visibility targets while 
preserving stable optimization behavior. During training, this 
SCR-guided reweighting reshapes gradient magnitudes according to 
target visibility; during inference, the model follows the exact same 
forward pass as the baseline, introducing no additional parameters or 
computational overhead.

REEM is implemented on top of the U-Net-based MSHNet 
framework~\cite{MSHNet2024} without modifying the network 
architecture, prediction head, or inference pipeline, and is readily 
applicable to other IRSTD architectures beyond MSHNet.

\begin{figure*}[t]
\FloatBarrier
\centering
\begin{tikzpicture}[
  font=\footnotesize,
  arrow/.style={-Latex, thick},
  block/.style={draw, rounded corners, align=center,
                minimum height=7mm, minimum width=27mm, inner sep=2.5pt},
  dashedblock/.style={draw, dashed, rounded corners, align=center,
                      minimum height=7mm, minimum width=31mm, inner sep=2.5pt},
  head/.style={draw, rounded corners, align=center,
               minimum height=6mm, minimum width=16mm, inner sep=1.5pt}
]

\node[block] (unet) {U-Net\\Backbone};

\node[right=18mm of unet] (ms_anchor) {};

\matrix (hm) [matrix of nodes,
  nodes={head},
  row sep=1.2mm,
  anchor=center] at (ms_anchor.center) {
  {Head $p_1$} \\
  {Head $p_2$} \\
  {Head $p_3$} \\
  {Head $p_4$} \\
};

\node[draw, rounded corners, fit=(hm), inner sep=3.5mm,
      label={[font=\footnotesize]above:Multi-Scale Head}] (msbox) {};

\node[block, right=8mm of msbox, minimum width=22mm] (fuse)
{Upsample\\Concatenate};

\node[block, right=10mm of fuse] (pred) {Final Prediction\\(Energy Map)};

\draw[arrow] (unet.east) -- (msbox.west);
\draw[arrow] (msbox.east) -- (fuse.west);
\draw[arrow] (fuse.east) -- (pred.west);

\node[dashedblock, above=8mm of pred] (sls) {SLS Loss};

\node[dashedblock, below=8mm of pred] (scr)
{GT-local SCR\\(from input image)\\$w(\hat{s})$};

\node[dashedblock, right=9mm of pred, minimum width=36mm] (total)
{Total Objective\\
$\mathcal{L}_{\mathrm{REEM}}
=\mathcal{L}_{\mathrm{SLS}}
-\lambda\,\mathbb{E}\!\left[w(\hat{s})\cdot \mathrm{sIoU}\right]$\\
(SCR-guided reweighting)};

\draw[arrow] (pred.north) -- (sls.south);
\draw[arrow] (pred.south) -- (scr.north);
\draw[arrow] (sls.east) -- (total.west);
\draw[arrow] (scr.east) -- (total.west);

\node[draw, dashed, rounded corners, inner sep=2.5mm,
      fit=(sls)(scr)(total)] {};
\node[font=\footnotesize] at ($(total.south)+(0,-6mm)$) {Training only};

\end{tikzpicture}
\caption{Overview of REEM integrated into MSHNet. REEM preserves the original MSHNet architecture and multi-scale prediction head.
During training, the standard SLS loss is applied to the predicted energy map,
while a GT-local SCR value is computed from the input image to derive a
differentiable weight $w(\hat{s})$, where $\hat{s}$ denotes the clipped SCR
defined in Eq.~\ref{eq:scr_weight}, which up-weights the learning signal for
low-visibility targets.
This SCR-guided weighting modifies only the optimization objective and introduces no additional parameters or inference-time overhead.
During inference, REEM follows the standard forward pass of the baseline model, and no SCR evaluation is performed.
REEM performs difficulty-aware optimization without altering the model architecture or inference behavior.}
\label{fig:reem_mshnet}
\end{figure*}
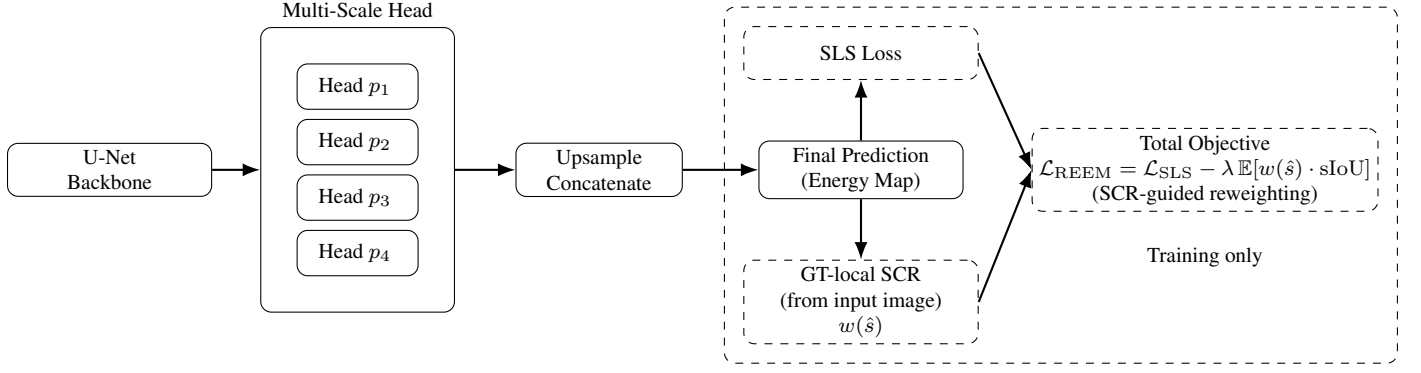

\subsection{Baseline Loss Functions for IRSTD}
\label{sec:baseline_loss}

Most deep learning-based IRSTD methods formulate the task as a pixel-wise segmentation problem and rely on overlap-driven objectives for supervision.
Among them, the Intersection-over-Union (IoU) loss is widely used to measure spatial agreement between predicted regions and ground-truth annotations.
However, IoU-based supervision is inherently insensitive to variations in target scale and location, which becomes particularly problematic in infrared small target detection where targets often occupy only a few pixels and exhibit weak spatial responses.

To alleviate these limitations, Liu \etal~\cite{MSHNet2024} introduced the \emph{Scale and Location Sensitive (SLS)} loss, which extends the IoU formulation by incorporating scale discrepancy and center displacement into the optimization objective.
The SLS loss decomposes supervision into scale-sensitive and location-sensitive components, encouraging the model to focus on small targets with inaccurate size or localization estimates.

Compared to conventional IoU-based objectives, SLS improves geometric sensitivity without introducing additional network parameters or architectural complexity, making it a strong baseline for IRSTD under small-target regimes.

\subsection{SCR-Guided Loss Reweighting}
\label{sec:scr_reward}

To address the limitations of purely geometry-driven supervision, we introduce 
a visibility-aware reweighting strategy based on the Signal-to-Clutter Ratio (SCR).
Instead of directly optimizing SCR or modifying the network architecture, REEM 
computes a ground-truth local SCR from the input image and uses it to modulate 
the learning signal through a differentiable weighting function, increasing the 
contribution of low-visibility targets while preserving the original geometric supervision.

Formally, let $\mathbf{I} \in \mathbb{R}^{H \times W}$ denote the input infrared image,
and let $\mathbf{G} \in \{0,1\}^{H \times W}$ denote the corresponding ground-truth target mask image of sizes $H$-by-$W$ pixels.
The set of target pixels is defined as
\begin{equation}
\mathcal{T} = \{ (x,y) \mid \mathbf{G}(x,y) = 1 \},
\end{equation}
and the local background region surrounding the target is defined as
\begin{equation}
\mathcal{B} = \{ (x,y) \mid (x,y) \in \mathcal{N}(\mathcal{T}) \land \mathbf{G}(x,y) = 0 \},
\end{equation}
where $\mathcal{N}(\mathcal{T})$ denotes a ring-shaped local 
neighborhood centered at the target bounding box. 
Specifically, the outer boundary is defined by scaling the 
bounding box dimensions by a factor of $r_{\mathrm{out}}=4.0$, 
and the inner boundary by $r_{\mathrm{in}}=1.5$. 
Target pixels are excluded from $\mathcal{B}$ to ensure that 
only background clutter statistics are captured.
When multiple targets are present in a single image, 
SCR is computed over the union of all target regions 
and their corresponding joint bounding box, yielding 
a single visibility weight per training sample.

The average intensity responses of the target and background regions are computed as
\begin{equation}
\mu_{\mathcal{T}} = \frac{1}{|\mathcal{T}|} \sum_{(x,y)\in \mathcal{T}} \mathbf{I}(x,y),
\end{equation}
\begin{equation}
\mu_{\mathcal{B}} = \frac{1}{|\mathcal{B}|} \sum_{(x,y)\in \mathcal{B}} \mathbf{I}(x,y),
\end{equation}
and the background standard deviation is
\begin{equation}
\sigma_{\mathcal{B}} =
\sqrt{\frac{1}{|\mathcal{B}|} \sum_{(x,y)\in \mathcal{B}}
\left(\mathbf{I}(x,y) - \mu_{\mathcal{B}}\right)^2 }.
\end{equation}

The local Signal-to-Clutter Ratio (SCR) is then defined as
\begin{equation}
s = \mathrm{SCR} =
\frac{|\mu_{\mathcal{T}} - \mu_{\mathcal{B}}|}{\sigma_{\mathcal{B}} + \epsilon},
\label{eq:scr_def}
\end{equation}
where $s$ denotes the local signal-to-clutter ratio and
$\epsilon$ is a small constant for numerical stability.

This formulation follows classical SCR definitions in infrared target
detection~\cite{Gao2013InfraredPatch}, and is computed locally from 
the input image using ground-truth supervision, serving as a visibility 
prior rather than a prediction-dependent objective.
Notably, the SLS loss~\cite{MSHNet2024} does not incorporate any
visibility-aware weighting; all samples contribute equally
to the geometric objective regardless of SCR level.
REEM therefore introduces a strictly complementary signal:
the SCR-guided term $w(\hat{s}) \cdot \mathrm{sIoU}$ modulates
only the soft-IoU component, preserving the original optimization
landscape while reshaping gradient magnitudes according to
target visibility.

\paragraph{Monotonic SCR weighting.}
To explicitly prioritize low-visibility targets, we define a bounded
monotonic weighting function as
\begin{equation}
\hat{s}=\operatorname{clip}(s,0,s_{\max}), \qquad
w(\hat{s}) = 1 + \alpha \frac{k}{\hat{s}+k},
\label{eq:scr_weight}
\end{equation}
where $s_{\max} = 12$ is a saturation threshold that caps extreme
high-contrast values to prevent outlier samples from dominating
the weighting, $k>0$ controls the saturation rate, and $\alpha>0$
determines the maximum upweighting level that separates difficult
low-SCR samples from high-visibility ones.
The function is strictly decreasing with respect to the clipped SCR:
\begin{equation}
\frac{\partial w}{\partial \hat{s}}
= -\alpha \frac{k}{(\hat{s}+k)^2} < 0,
\end{equation}
and satisfies $w\in[1,1+\alpha]$ for $\hat{s}\ge0$.
This formulation ensures that lower-SCR targets receive higher
optimization emphasis while maintaining bounded and differentiable
gradients. The weighting is applied only during training and introduces
no additional inference-time computation.

The total training objective is defined as:
\begin{equation}
\mathcal{L}_{\text{REEM}}
= \mathcal{L}_{\text{SLS}}
- \lambda \, \mathbb{E}\!\left[ w(\hat{s}) \cdot \mathrm{sIoU} \right],
\end{equation}
where $\mathrm{sIoU}$ denotes the soft Intersection-over-Union
component of the SLS loss~\cite{MSHNet2024}, and
$\lambda>0$ controls the strength of SCR-guided objective modulation.

Unlike heuristic reweighting strategies, $w(\hat{s})$ is computed 
solely from ground-truth statistics and remains independent of model 
predictions, introducing no handcrafted decision rules, auxiliary 
optimization targets, or inference-time overhead.
\FloatBarrier
\section{Experimental Analysis}
\label{sec:experiments}
We evaluate REEM on two standard IRSTD benchmarks and compare 
against both traditional and deep learning-based methods. 
We also analyze performance across SCR bins to verify that 
the proposed reweighting strategy specifically benefits 
low-visibility targets.
\subsection{Experimental Setup}
\label{sec:exp_setup}

\begin{table*}[t]
\centering
\caption{Overall performance comparison on IRSTD-1k and NUDT-SIRST datasets.}
\label{tab:overall_comparison}
\small
\begin{tabular}{|l|l|c|c|c|c|c|c|}
\hline
Method & Description &
\multicolumn{3}{c|}{IRSTD-1k} & \multicolumn{3}{c|}{NUDT-SIRST} \\
\cline{3-8}
 &  & IoU$\uparrow$ & Pd$\uparrow$ & FA$\downarrow$ & IoU$\uparrow$ & Pd$\uparrow$ & FA$\downarrow$ \\
\hline
Top-Hat (Morphological)~\cite{tophat} & Filtering
& 10.06 & 75.11 & 1432 & 20.72 & 78.41 & 166.7 \\

Max-Median~\cite{Deshpande1999MaxMedian} & Filtering
& 6.998 & 65.21 & 59.73 & 4.197 & 58.41 & 36.89 \\

WSLCM~\cite{Han2020WSLCM} & Local Contrast
& 3.452 & 72.44 & 6619 & 2.283 & 56.82 & 1309 \\

TLLCM~\cite{Han2019TriLayerLCM} & Local Contrast
& 3.311 & 77.39 & 6738 & 2.176 & 62.01 & 1608 \\

IPI~\cite{Gao2013InfraredPatch} & Low Rank
& 27.92 & 81.37 & 16.18 & 17.76 & 74.49 & 41.23 \\

NRAM~\cite{Zhang2018NRAM} & Low Rank
& 15.25 & 70.68 & 16.93 & 6.927 & 56.40 & 19.27 \\

RIPT~\cite{Dai2017RIPT} & Low Rank
& 14.11 & 77.55 & 28.31 & 29.44 & 91.85 & 344.3 \\

PSTNN~\cite{Zhu2020TNLRS} & Low Rank
& 24.57 & 71.99 & 35.26 & 14.85 & 66.13 & 44.17 \\

MSLSTIPT~\cite{Yang2020Subspace} & Low Rank
& 11.43 & 79.03 & 1524 & 8.342 & 47.40 & 888.1 \\
\hline
MDvsFA~\cite{Wang2019MDvsFA} & Deep Learning
& 37.34 & 83.71 & 88.52 & 35.86 & 85.22 & 95.37 \\

ALCNet~\cite{Dai2021ALCNet} & Deep Learning
& 65.68 & 89.25 & 27.71 & 72.89 & 96.19 & 30.40 \\

ISNet~\cite{Zhang2022ISNet} & Deep Learning
& 62.88 & 92.59 & 27.92 & 67.86 & 92.59 & 34.65 \\

DNANet~\cite{DNANet2023} & Deep Learning
& 65.35 & 86.05 & 21.93 & 78.28 & 93.33 & 15.01 \\

MSHNet~\cite{MSHNet2024} & Deep Learning
& 65.60 & 93.20 & 13.51 & 74.52 & 95.37 & 29.00 \\
\hline
\textbf{REEM (ours)} & \textbf{SCR-Aware Reweighting}
& \textbf{68.44} & \textbf{93.88} & \textbf{6.30}
& \textbf{79.86} & \textbf{97.52} & \textbf{11.21} \\
\hline
\end{tabular}
\vspace{2mm}
\begin{minipage}{0.98\linewidth}
\footnotesize
\textbf{Note:}
Higher IoU and Pd indicate better performance, while lower FA is preferred. Results of existing methods are taken directly from the original MSHNet paper, except MSHNet and DNANet which are reproduced using their official implementations under identical experimental settings. All methods are evaluated using a fixed threshold $\tau=0.5$; Pd and FA are computed with image-averaged normalization (ppm) unless otherwise stated.
\end{minipage}
\end{table*}

\paragraph{Datasets.}
All experiments are conducted on two widely used infrared small target detection
benchmarks, IRSTD-1k~\cite{Zhang2022ISNet} and NUDT-SIRST~\cite{DNANet2023},
which contain 1,001 and 1,327 infrared images, respectively.
In IRSTD-1k, targets occupy a median area of 45\,px on 512$\times$512 images;
in NUDT-SIRST, the median target area is 40\,px on 256$\times$256 images,
confirming that both datasets reside firmly in the small-target regime.
We follow the predefined training and testing partitions provided with each dataset
to ensure direct comparability with prior IRSTD methods.

\paragraph{Evaluation metrics.}
We adopt standard evaluation metrics for infrared small target detection, including Intersection-over-Union (IoU), Probability of Detection (Pd), and False Alarm rate (FA).
IoU measures pixel-level overlap between predictions and ground truth, Pd evaluates target-level detection accuracy, and FA quantifies the number of false alarm pixels normalized by the image area.
Higher IoU and Pd values indicate better detection performance, while lower FA is preferred.
Table~\ref{tab:overall_comparison} presents the overall performance comparison on the IRSTD-1k and NUDT-SIRST datasets.
Results of existing methods are taken directly from the original MSHNet paper for fair comparison.
Unlike prior approaches that rely on increasingly complex network architectures, REEM modifies only the optimization objective and introduces no additional parameters or inference-time overhead.

\paragraph{Evaluation protocol.}
Unless otherwise stated, we report fixed-threshold results at $\tau=0.5$.
For Pd, we follow the common target-level definition (image-averaged unless specified), and FA is reported in parts-per-million (ppm) normalized by the image area.

\paragraph{Implementation details.}
REEM is implemented on top of the publicly released MSHNet framework~\cite{MSHNet2024} without modifying the original network architecture.
The backbone follows the U-Net-based design with a multi-scale prediction head.
Input images are resized to $256\times256$.
Training is performed using the AdaGrad optimizer with an initial learning rate of 0.05, following the original MSHNet settings.
Unless otherwise stated, all data preprocessing steps, training schedules,
and testing pipelines strictly follow the official MSHNet implementation
to ensure that performance differences arise solely from the proposed
SCR-guided loss reweighting mechanism, rather than architectural or
implementation-related factors.

\begin{figure*}[t]
\centering
\setlength{\tabcolsep}{2pt}

\begin{minipage}{0.16\textwidth}\centering\footnotesize\textbf{Infrared Image}\end{minipage}
\begin{minipage}{0.16\textwidth}\centering\footnotesize\textbf{Ground Truth}\end{minipage}
\begin{minipage}{0.16\textwidth}\centering\footnotesize\textbf{Top-Hat~\cite{tophat}}\end{minipage}
\begin{minipage}{0.16\textwidth}\centering\footnotesize\textbf{DNANet~\cite{DNANet2023}}\end{minipage}
\begin{minipage}{0.16\textwidth}\centering\footnotesize\textbf{MSHNet~\cite{MSHNet2024}}\end{minipage}
\begin{minipage}{0.16\textwidth}\centering\footnotesize\textbf{REEM (Ours)}\end{minipage}

\vspace{2pt}

\begin{subfigure}[t]{0.16\textwidth}\centering
\includegraphics[width=\linewidth]{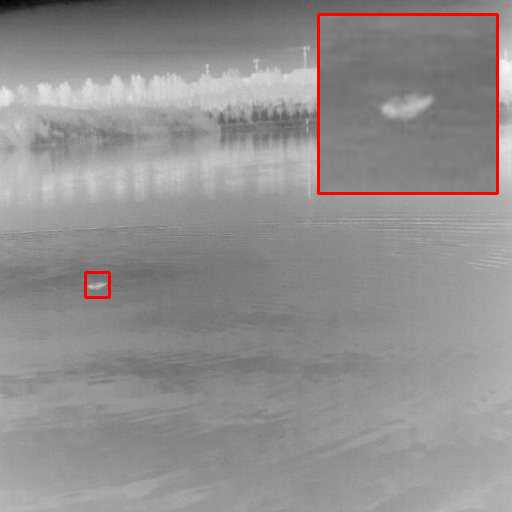}
\end{subfigure}
\begin{subfigure}[t]{0.16\textwidth}\centering
\includegraphics[width=\linewidth]{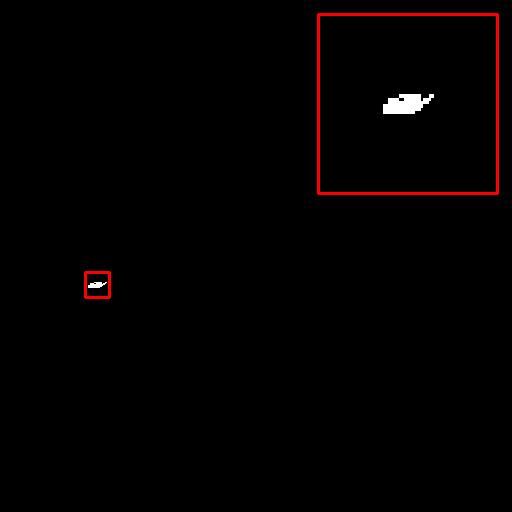}
\end{subfigure}
\begin{subfigure}[t]{0.16\textwidth}\centering
\includegraphics[width=\linewidth]{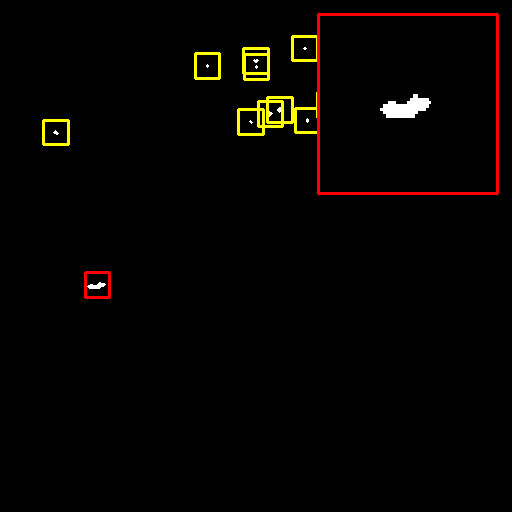}
\end{subfigure}
\begin{subfigure}[t]{0.16\textwidth}\centering
\includegraphics[width=\linewidth]{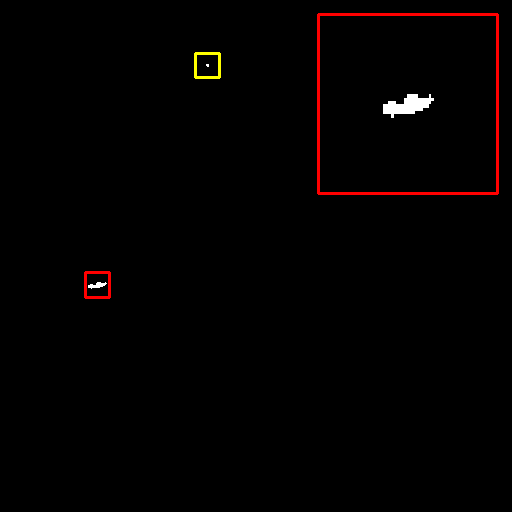}
\end{subfigure}
\begin{subfigure}[t]{0.16\textwidth}\centering
\includegraphics[width=\linewidth]{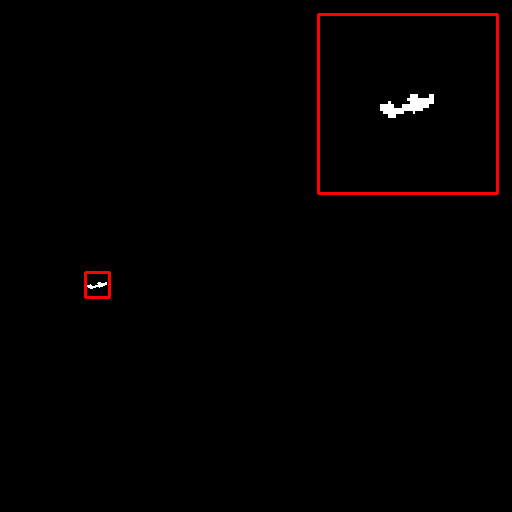}
\end{subfigure}
\begin{subfigure}[t]{0.16\textwidth}\centering
\includegraphics[width=\linewidth]{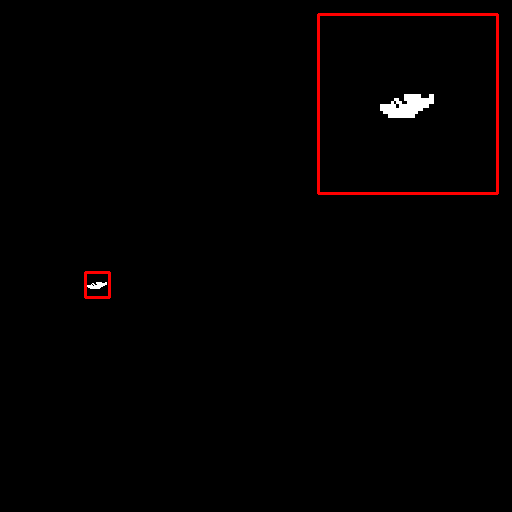}
\end{subfigure}

\vspace{3pt}

\begin{subfigure}[t]{0.16\textwidth}\centering
\includegraphics[width=\linewidth]{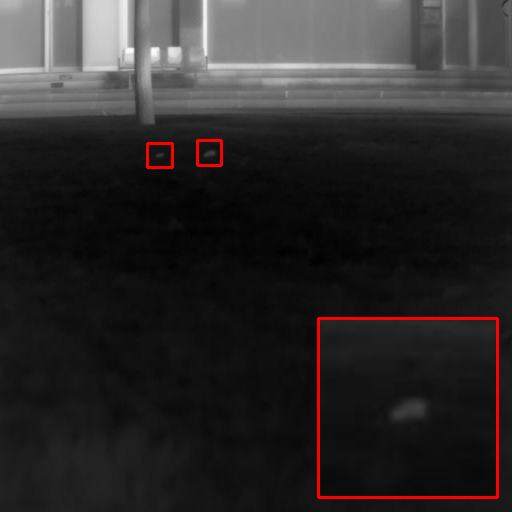}
\end{subfigure}
\begin{subfigure}[t]{0.16\textwidth}\centering
\includegraphics[width=\linewidth]{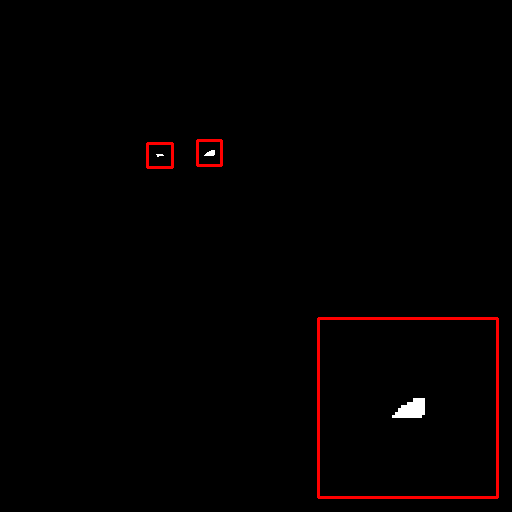}
\end{subfigure}
\begin{subfigure}[t]{0.16\textwidth}\centering
\includegraphics[width=\linewidth]{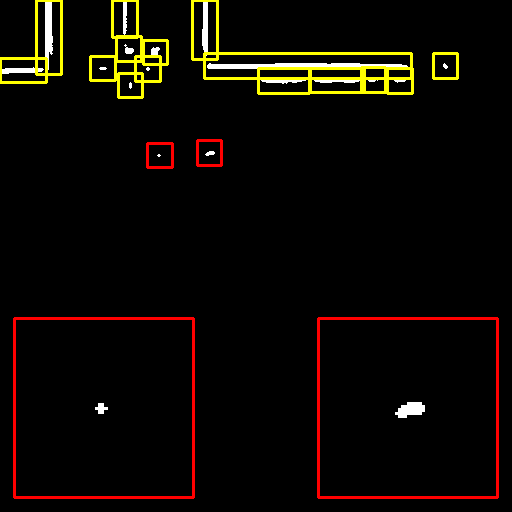}
\end{subfigure}
\begin{subfigure}[t]{0.16\textwidth}\centering
\includegraphics[width=\linewidth]{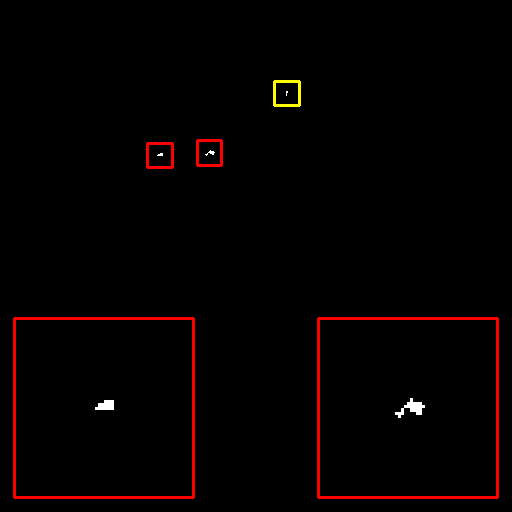}
\end{subfigure}
\begin{subfigure}[t]{0.16\textwidth}\centering
\includegraphics[width=\linewidth]{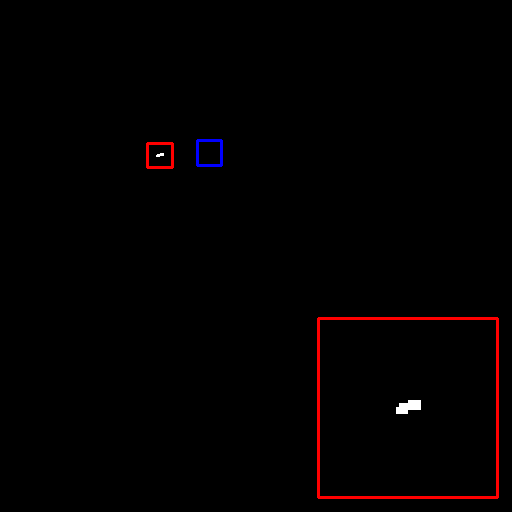}
\end{subfigure}
\begin{subfigure}[t]{0.16\textwidth}\centering
\includegraphics[width=\linewidth]{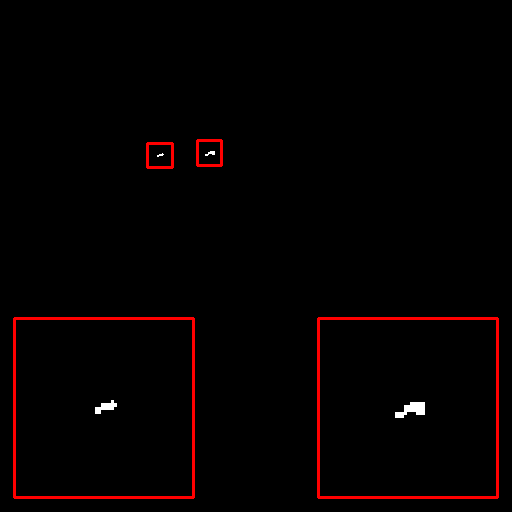}
\end{subfigure}

\vspace{3pt}

\begin{subfigure}[t]{0.16\textwidth}\centering
\includegraphics[width=\linewidth]{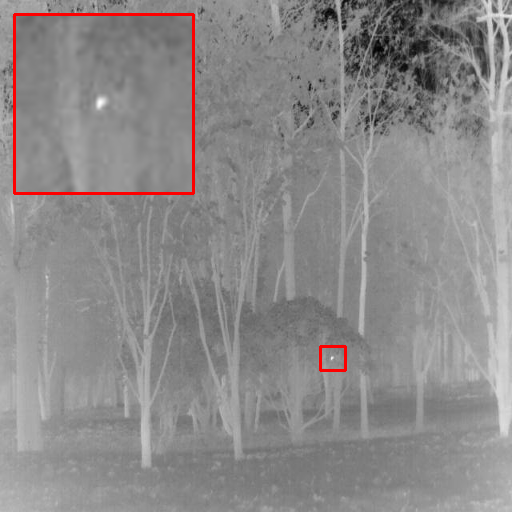}
\end{subfigure}
\begin{subfigure}[t]{0.16\textwidth}\centering
\includegraphics[width=\linewidth]{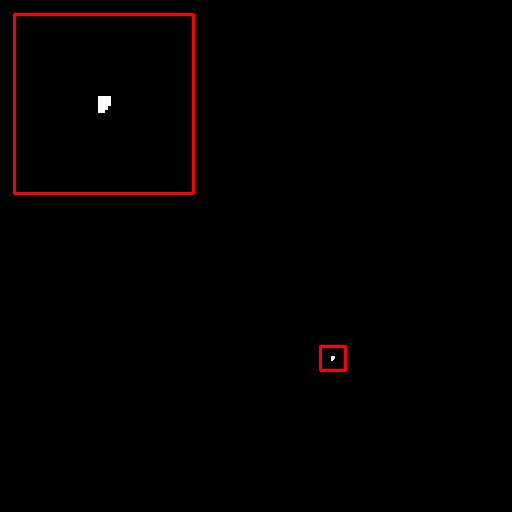}
\end{subfigure}
\begin{subfigure}[t]{0.16\textwidth}\centering
\includegraphics[width=\linewidth]{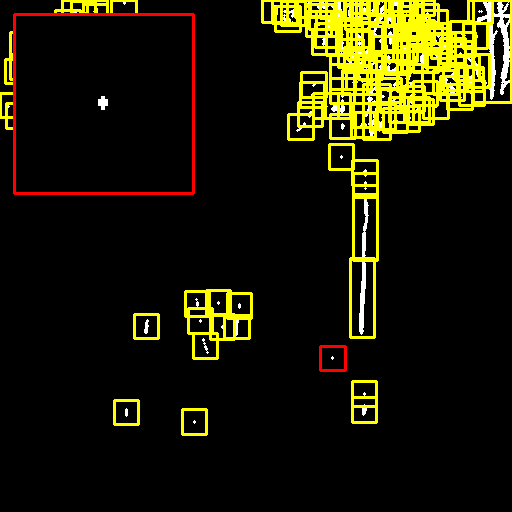}
\end{subfigure}
\begin{subfigure}[t]{0.16\textwidth}\centering
\includegraphics[width=\linewidth]{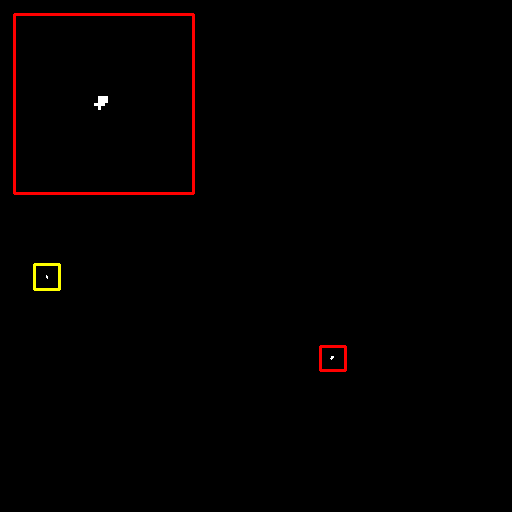}
\end{subfigure}
\begin{subfigure}[t]{0.16\textwidth}\centering
\includegraphics[width=\linewidth]{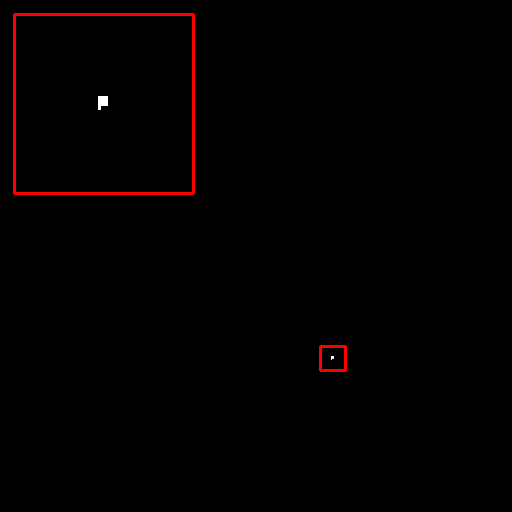}
\end{subfigure}
\begin{subfigure}[t]{0.16\textwidth}\centering
\includegraphics[width=\linewidth]{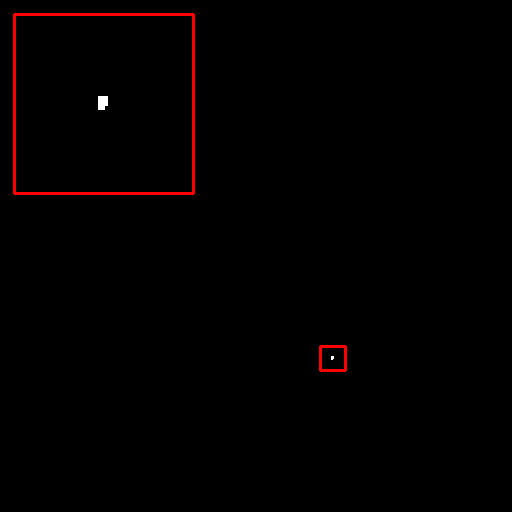}
\end{subfigure}

\vspace{3pt}

\begin{subfigure}[t]{0.16\textwidth}\centering
\includegraphics[width=\linewidth]{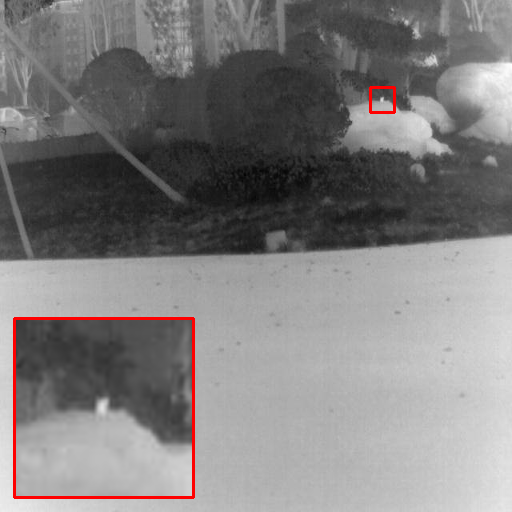}
\end{subfigure}
\begin{subfigure}[t]{0.16\textwidth}\centering
\includegraphics[width=\linewidth]{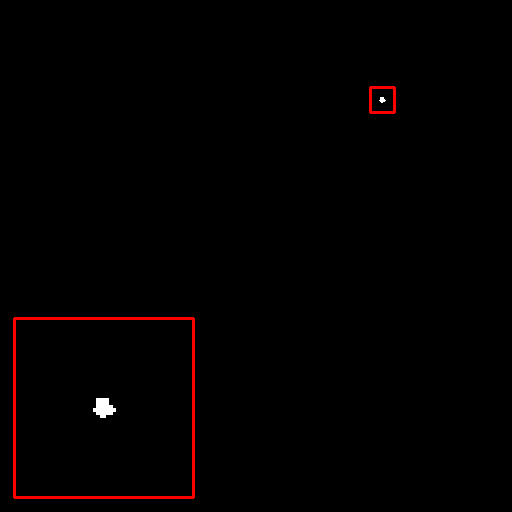}
\end{subfigure}
\begin{subfigure}[t]{0.16\textwidth}\centering
\includegraphics[width=\linewidth]{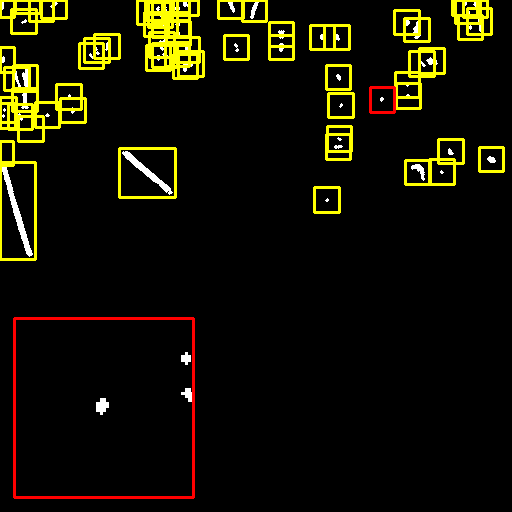}
\end{subfigure}
\begin{subfigure}[t]{0.16\textwidth}\centering
\includegraphics[width=\linewidth]{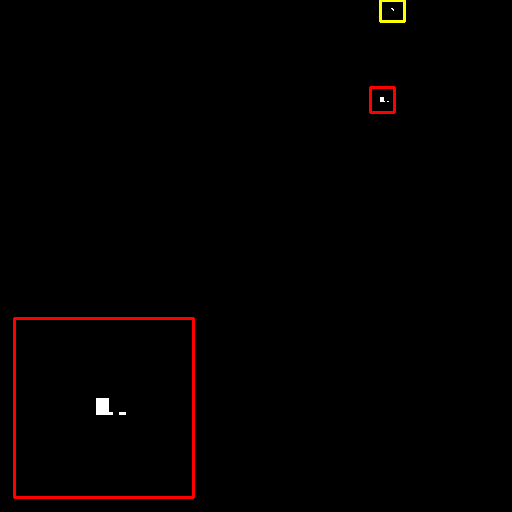}
\end{subfigure}
\begin{subfigure}[t]{0.16\textwidth}\centering
\includegraphics[width=\linewidth]{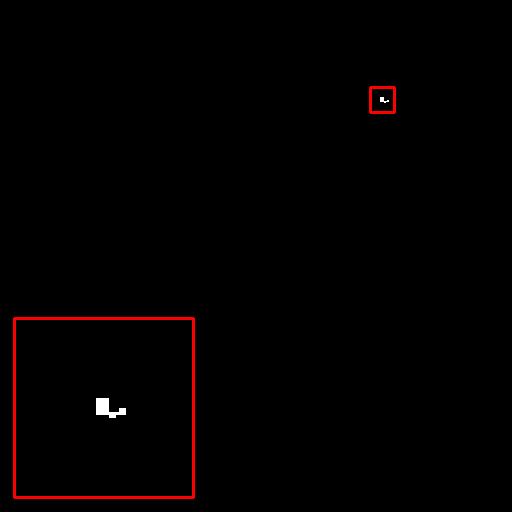}
\end{subfigure}
\begin{subfigure}[t]{0.16\textwidth}\centering
\includegraphics[width=\linewidth]{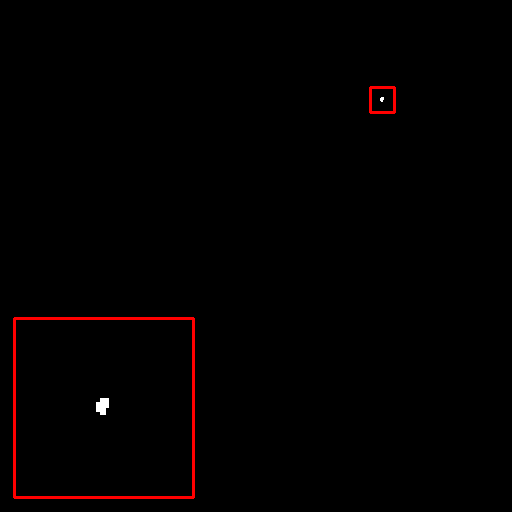}
\end{subfigure}


\caption{
Visual comparison of detection results on representative infrared images.
Correct detections, missed targets, and false alarms are indicated by red, blue,
and yellow bounding boxes, respectively.
Zoomed-in views of target regions are provided in the image corners for clarity.
Compared to prior methods, REEM produces cleaner energy responses by suppressing
clutter-induced activations while preserving weak small-target signals.
}

\label{fig:vis_multi_xdu_clean}
\end{figure*}

\paragraph{Key observation.}
REEM primarily improves \emph{reliability} by suppressing 
clutter-induced responses in low-visibility cases.
Under fixed-threshold evaluation ($\tau=0.5$) on IRSTD-1k, REEM 
surpasses the baseline MSHNet while preserving the same inference 
architecture: IoU improves from 65.60\% to 68.44\%, FA decreases 
from 13.51\,ppm to 6.30\,ppm, with a simultaneous gain in Pd 
(93.20\% to 93.88\%). On NUDT-SIRST, consistent improvements are 
observed across all metrics, with IoU increasing from 74.52\% to 
79.86\% and FA decreasing from 29.00\,ppm to 11.21\,ppm.
In SCR-imbalanced datasets such as IRSTD-1k, where approximately 
11\% of samples fall below $\mathrm{SCR}<2$, standard overlap-driven 
objectives tend to overfit to easy high-SCR samples.
SCR-guided reweighting instead increases the contribution of hard 
samples, improving target--background separability and reducing 
false alarms without inference-time heuristics.
As summarized in Table~\ref{tab:scr_bins_combined}, performance 
gains are strongest in low- and mid-SCR regimes: on IRSTD-1k, FA 
drops from 18.53\,ppm to 3.27\,ppm in the $[1,2)$ bin, and from 
33.21\,ppm to 17.95\,ppm in the $[2,4)$ bin. On NUDT-SIRST, IoU 
improves by more than 9 points in the $\mathrm{SCR}<1$ regime 
(67.96\% $\rightarrow$ 77.16\%). Differences become marginal at 
$\mathrm{SCR}\geq8$ where detection is already near saturation, 
consistent with the expected behavior of the proposed weighting 
function.

\paragraph{Failure cases.}
REEM exhibits a consistent limitation in the high-SCR regime
($\mathrm{SCR}\geq8$), where minor degradation in FA is observed
on both datasets (IRSTD-1k: 6.36$\rightarrow$6.54\,ppm;
NUDT-SIRST: 4.44$\rightarrow$9.58\,ppm).
This behavior stems from the weighting function $w(\hat{s})$,
which approaches but never reaches $1$ for finite SCR values,
resulting in residual gradient updates on already well-detected
high-visibility targets.
Under a fixed decision threshold, these unnecessary updates
slightly shift the decision boundary and can introduce spurious
activations on high-contrast background regions.
Since REEM is designed to prioritize low-SCR targets, this
trade-off is an expected consequence of the proposed objective
design rather than a fundamental failure of the method.
A straightforward mitigation would be to disable REEM's 
reweighting entirely for high-visibility targets; we leave 
this as future work.

\begin{table*}[!t]
\centering
\small
{\setlength{\tabcolsep}{3pt}
\begin{tabular}{c c ccc ccc}
\hline
Dataset &
SCR &
\multicolumn{3}{c}{\textbf{BASE (MSHNet)}~\cite{MSHNet2024}} &
\multicolumn{3}{c}{\textbf{REEM}} \\
\cline{3-8}
 &  & IoU$\uparrow$ & Pd$\uparrow$ & FA$\downarrow$ &
          IoU$\uparrow$ & Pd$\uparrow$ & FA$\downarrow$ \\
\hline
\multirow{5}{*}{\textbf{IRSTD-1k}}
& $<1$
& 45.06 & 78.12 & 13.35
& \textbf{45.65} & \textbf{84.37} & \textbf{3.81} \\

& $[1,2)$
& 43.87 & 69.05 & 18.53
& \textbf{48.08} & \textbf{72.62} & \textbf{3.27} \\

& $[2,4)$
& 59.50 & \textbf{89.22} & 33.21
& \textbf{61.51} & 88.73 & \textbf{17.95} \\

& $[4,8)$
& 62.43 & 96.17 & 13.51
& \textbf{64.00} & \textbf{96.99} & \textbf{3.00} \\

& $\geq 8$
& \textbf{63.08} & \textbf{99.60} & \textbf{6.36}
& 62.07 & 98.41 & 6.54 \\
\hline

\multirow{5}{*}{\textbf{NUDT-SIRST}}
& $<1$
& 67.96 & 87.99 & 39.04
& \textbf{77.16} & \textbf{94.16} & \textbf{22.19} \\

& $[1,2)$
& 67.53 & 91.28 & 40.46
& \textbf{75.14} & \textbf{95.41} & \textbf{11.90} \\

& $[2,4)$
& 75.16 & 94.55 & 55.58
& \textbf{79.45} & \textbf{97.58} & \textbf{12.95} \\

& $[4,8)$
& 79.97 & 97.87 & 13.53
& \textbf{82.33} & \textbf{98.58} & \textbf{4.65} \\

& $\geq 8$
& 86.47 & \textbf{100.00} & \textbf{4.44}
& \textbf{89.77} & 99.42 & 9.58 \\
\hline
\end{tabular}
}
\caption{SCR-binned performance on IRSTD-1k and NUDT-SIRST test sets at a fixed threshold $\tau=0.5$.
SCR is computed using local target--background statistics with edges $\{1,2,4,8\}$.
All values are reported as percentages (\%) except FA (ppm).
Best results within each SCR bin are highlighted in bold.}
\label{tab:scr_bins_combined}
\end{table*}
\FloatBarrier

\paragraph{Inference efficiency.}
Since REEM modifies only the training objective and introduces no additional
computations during inference, its runtime performance remains identical to the
baseline MSHNet in terms of architectural complexity.
On an NVIDIA RTX~5080 GPU, the PyTorch implementation achieves
approximately \textbf{131.8 FPS} ($\sim$7.6\,ms per frame) for
$256\times256$ inputs.
After TensorRT FP16 deployment, throughput increases to
\textbf{694.9 FPS} ($\sim$1.44\,ms per frame), corresponding to a
\textbf{5.3$\times$ speedup}.
Importantly, this acceleration does not degrade detection quality:
the IoU changes only marginally from $68.49\%$ (PyTorch) to
$68.23\%$ (TensorRT), which is within normal numerical variance.
These results confirm that the proposed SCR-guided loss reweighting
incurs \emph{no inference-time overhead} while remaining fully
compatible with high-throughput deployment pipelines.
\balance
\section{Conclusion}
\label{sec:conclusion}

This paper introduced REEM, a physically grounded SCR-guided loss reweighting
strategy for infrared small target detection.
Rather than designing new network architectures, REEM improves optimization by
incorporating signal-to-clutter information as a visibility-aware modulation of
the training objective.
This formulation enables difficulty-aware learning that emphasizes low-visibility
targets while preserving the original model architecture, training pipeline,
and inference-time behavior.

Extensive experiments on IRSTD-1k and NUDT-SIRST demonstrate that REEM achieves
consistent performance improvements over the MSHNet baseline, with particularly
strong gains in low-SCR regimes where conventional overlap-driven supervision
often struggles.
The results show that integrating physically meaningful signal characteristics
into the optimization process provides an effective and lightweight complement
to geometry-based loss formulations, improving robustness without introducing
additional parameters or computational overhead.

Future work will explore extending SCR-guided optimization 
to multi-frame and temporal infrared detection settings, 
investigating signal-aware alternatives to generic 
difficulty-aware strategies such as focal loss or hard 
example mining, as well as broader signal-aware supervision 
strategies for challenging low-contrast vision tasks.

\section*{Acknowledgements}
Beh\c{c}et U\u{g}ur T\"{o}reyin's work was supported in part by 
the Scientific and Technological Research Council of Turkiye 
(TUBITAK) with 1515 Frontier Research and Development Laboratories 
Support Program for the BTS Advanced AI Hub: BTS Autonomous Networks 
and Data Innovation Laboratory under Project 5239903; in part by the 
Turk Telekom 6G Research and Development Laboratory under Project 
5249902; and in part by the Scientific Research Projects Coordination 
Department (BAP), Istanbul Technical University under Project 
ITU-BAP MGA-2024-45372.

{
    \small
    \bibliographystyle{ieeenat_fullname}
    \bibliography{main}
}


\end{document}